\documentclass[pdflatex,sn-chicago]{sn-jnl}

\usepackage{floatrow}
\usepackage[label font=bf, 
            labelformat=simple]{subfig}

\usepackage{longtable}
\usepackage{graphicx}%
\usepackage{multirow}%
\usepackage{amsmath,amssymb,amsfonts}%
\usepackage{amsthm}%
\usepackage{mathrsfs}%
\usepackage[title]{appendix}%
\usepackage{xcolor}%
\usepackage{textcomp}%
\usepackage{manyfoot}%
\usepackage{booktabs}%
\usepackage{algorithm}%
\usepackage{algorithmicx}%
\usepackage{algpseudocode}%
\usepackage{listings}%

\theoremstyle{thmstyleone}%
\theoremstyle{thmstyletwo}%

\theoremstyle{thmstylethree}%

\begin{document}

\title[An Expert-grounded benchmark of General Purpose LLMs in LCA]{An Expert-grounded benchmark of General Purpose LLMs in LCA}


\author*[1]{\fnm{Artur} \sur{Donaldson}}\email{donaldson@pre-sustainability.com}

\author[2]{\fnm{Bharathan} \sur{Balaji}}
\equalcont{}

\author[3]{\fnm{Cajetan} \sur{Oriekezie}}
\equalcont{}

\author[4]{\fnm{Manish} \sur{Kumar}}
\equalcont{}

\author[5]{\fnm{Laure} \sur{Patouillard}}
\equalcont{}

\affil*[1]{\orgname{PRé Sustainability B.V.}, \orgaddress{\street{Stationsplein 121}, \city{Amersfoort}, \postcode{3818 LE}, \country{The Netherlands}}}

\affil[2]{
\orgname{Amazon}, \orgaddress{
\city{Seattle}, 
\state{Washington}, \country{United States}}}

\affil[3]{\orgdiv{School of Engineering}, \orgname{London South Bank University}, \orgaddress{103 Borough Road}, 
\city{London},
\postcode{SE1 0AA}, 
\country{United Kingdom}}

\affil[4]{\orgdiv{Helmholtz Institute Ulm-Electrochemical Energy Storage (HIU) }, \orgaddress{{Helmholtzstraße 11}, 
\city{Ulm}, \postcode{89081}, Karlsruhe Institute of Technology, \country{Germany}}}

\affil[5]{\orgdiv{CIRAIG, Department of Chemical Engineering}, \orgname{Polytechnique Montréal}, \orgaddress{{3333 Queen Mary Rd Suite 310},
\city{Montreal}, 
\postcode{H3V 1A2}
\state{QC}, 
\country{Canada}}}

\abstract{\textbf{Purpose:} Artificial intelligence (AI), and in particular large language models (LLMs), are increasingly being explored as tools to support life cycle assessment (LCA). While demonstrations exist across environmental and social domains, systematic evidence on their reliability, robustness, and usability remains limited. This study provides the first expert-grounded benchmark of LLMs in LCA, addressing the absence of standardized evaluation frameworks in a field where no clear ground truth or consensus protocols exist.

\textbf{Methods:} We evaluated eleven general-purpose LLMs, spanning both commercial and open-source families, across 22 LCA-related tasks. Seventeen experienced practitioners reviewed model outputs against criteria directly relevant to LCA practice, including scientific accuracy, explanation quality, robustness, verifiability, and adherence to instructions. We collected 168 expert reviews.

\textbf{Results:} Experts judged 37\% of responses to contain inaccurate or misleading information. Ratings of accuracy and quality of explanation were generally rated ``average'' or ``good'' on many models even smaller models, and format adherence was generally rated favourably. Hallucination rates varied significantly, with some models producing hallucinated citations at rates of up to 40\%. There was no clear-cut distinction between ratings on open-weight versus closed-weight LLMs, with open-weight models outperforming or competing on par with closed-weight models on criteria such as accuracy and quality of explanation. 

\textbf{Conclusion:} These findings highlight the risks of applying LLMs naïvely in LCA, such as when LLMs are treated as free-form oracles, while also showing benefits especially around quality of explanation and alleviating labour intensiveness of simple tasks. The use of general-purpose LLMs without grounding mechanisms presents quantifiable risks that can directly affect LCA result quality, such as hallucination of citations. This work underscore the need for larger, more diverse benchmarks to better quantify and improve standards around LLM use, including grounding mechanisms to lower the rate of occurrence of undesirable model behaviour. While no system can eliminate errors, development of robust testing standards offer clarity, and a pathway to improve and validate best practices when LLMs are used in LCA workflows.
}




\maketitle

\section{Introduction}\label{sec1}

Artificial intelligence (AI), and in particular large language models (LLMs), are increasingly being explored as tools to support different phases of the life cycle assessment methodology. Recent work has shown that LLMs have the potential to accelerate data-intensive tasks such as product carbon footprinting. For example, AutoPCF demonstrates how cradle-to-gate inventories can be automatically generated in minutes rather than days, using a combination of LLM reasoning and structured datasets~\cite{luo2023autopcf}. Domain-specific applications are also emerging. For instance, \cite{turhan2023life} developed an LLM-enabled tool to support LCA of bio-based construction materials, while \cite{chen2024advancing} illustrated the use of customized LLM pipelines for assessing green hydrogen supply chains. These examples highlight the potential for AI to reduce manual labour and expand the scalability of LCA practice.

More recent efforts focus on retrieval-augmented generation (RAG) for emission factor (EF) selection. Parakeet operationalizes RAG by embedding product and process descriptions, retrieving candidate factors from curated repositories, ranking and explaining them, and keeping a human in the loop~\cite{balaji2025emission}. Reported benefits include improved transparency and reviewer throughput when EF recommendations are paired with expert adjudication, articulating a pragmatic division of labor between experts and models. Other studies have extended RAG pipelines with general-purpose LLMs, enabling near real-time product carbon footprint updates across multiple industries~\cite{zhang2024carbonreveal,wang2024carbon}. These developments provide a template for LCA deployments in which LLMs act as controllable language engines grounded in vetted corpora rather than as free-form oracles.

Beyond environmental LCA, early studies of LLMs in social life cycle assessment (S-LCA) point to opportunities for automating qualitative assessments.~\cite{cole2025towards} evaluated how LLMs can support the analysis of labour practices, community impacts, and human rights in S-LCA, demonstrating efficiency gains while benchmarking results against expert reviews. This complements survey work highlighting both rising interest and persistent concerns in the LCA community, particularly around transparency, reproducibility, and credibility~\cite{goridkov2024s}. More broadly, analyses of LLM behaviour in other domains have underscored risks related to hallucination, unreliable citations, and difficulties in instruction following~\cite{ji2023survey, khourdajie_role_2025}, which raise important questions about the responsible deployment of LLMs in LCA. Hallucination in particular has been consistently flagged as a systemic problem in generative AI, with some authors arguing that hallucinations persist because standard training and evaluation practices reward guessing and penalise uncertainty. Under such grading, a model that truthfully abstains can score worse than an otherwise similar model that always guesses, creating systemic disincentives to admit not knowing. \cite{kalai_why_2025}.

The LCA community has begun to map this emerging landscape.~\cite{popowicz2025digital} situate AI within broader digital transformations, arguing that value arises from task-specific orchestration rather than generic chat.~\cite{preuss2024large} survey opportunities and risks of LLMs across LCA phases, emphasizing the heterogeneity of tasks—from goal and scope definition to EF mapping and interpretation—and the need for evaluation harnesses tuned to each. Likewise, \cite{macmaster2024testing} probe targeted uses such as LLM-assisted data-quality assessment of secondary datasets, finding productivity gains under template-guided prompting but reaffirming the need for expert oversight~\cite{macmaster2024testing}. Together, these papers converge on the stance that scaffolded AI is more promising than unconstrained generative use.

A central challenge in benchmarking LLM outputs for LCA is that the field itself lacks a well-defined ground truth. Replication studies often yield widely varying results due to subjective methodological choices like boundary settings and allocation rules — a phenomenon well-documented in LCA literature~\cite{tan2025uncertainty}. For instance, estimates of bioenergy emissions range from -56g to 163g of CO2eq/MJ depending on system boundaries and soil assumptions~\cite{mcmanus2015challenge}. Moreover, there are no widely accepted scoring mechanisms or consensus protocols for adjudicating expert agreement in LCA~\cite{edelen2018creation}. These conditions make it difficult to determine when LLM-generated answers are “correct” or “good enough.” Yet, benchmarks are precisely what enable scientific progress and trust: they must be methodologically robust, large and diverse enough to reflect real-world variability, and explicitly capture expert disagreement. To tackle this issue, we draw inspiration from domains beyond Life Cycle Assessment where expert-grounded benchmark datasets are well established for assessing reasoning of LLM performance, including in ``hard'' sciences~\cite{rein_gpqa_2024}, law ~\cite{guha_legalbench_2023} and bioinformatics~\cite{mitchener_bixbench_2025}, and aim to establish a first benchmark for evaluating reasoning of LLMs in the field of life cycle assessment.

Despite growing interest, today’s evidence base remains limited to single-system demonstrations and narrative case studies. What is missing is a comparative, expert-grounded evaluation of how contemporary general-purpose LLMs — spanning both commercial and open-source families —perform across representative LCA tasks when used as practitioners actually use them (i.e., baseline prompting, without proprietary UI augmentations). This paper addresses these gaps by presenting the first expert-grounded benchmark of LLMs in LCA. Seventeen experienced practitioners reviewed 168 AI-generated answers from eleven general-purpose LLMs, assessing scientific accuracy, explanation quality, robustness, reproducibility, and adherence to instructions. By systematically comparing models under these dimensions, we provide empirical insights into current strengths and limitations of LLMs for LCA, and lay foundational guidance toward developing trusted benchmarking frameworks for future AI integration in the LCA field.

\section{Methodology}\label{sec2}
The methodology was designed to systematically evaluate how LLMs perform on tasks relevant to life cycle assessment (LCA), while grounding the evaluation in expert judgment from experienced LCA practitioners. Our approach builds on contemporary work with the United National Environment Programme (UNEP) - Life Cycle Initiative Working Group on AI and LCA, including surveys on ethical values and current AI tool use in LCA. The insights were mapped to representative LCA task types that could be formulated as prompts for LLMs, ensuring that the benchmark reflects real-world practice rather than purely academic exercises.

The overall evaluation process combined four complementary stages: (i) initial survey design, (ii) generating model responses across a diverse set of LCA tasks using multiple LLMs, and (iii) engaging expert reviewers to assess these responses against agreed evaluation criteria, (iv) result analysis. This design allowed us to capture both technical performance and practitioner perspectives, which are essential for building trust in the use of AI in LCA. 

\begin{figure}
    \centering
    \includegraphics[width=1\linewidth]{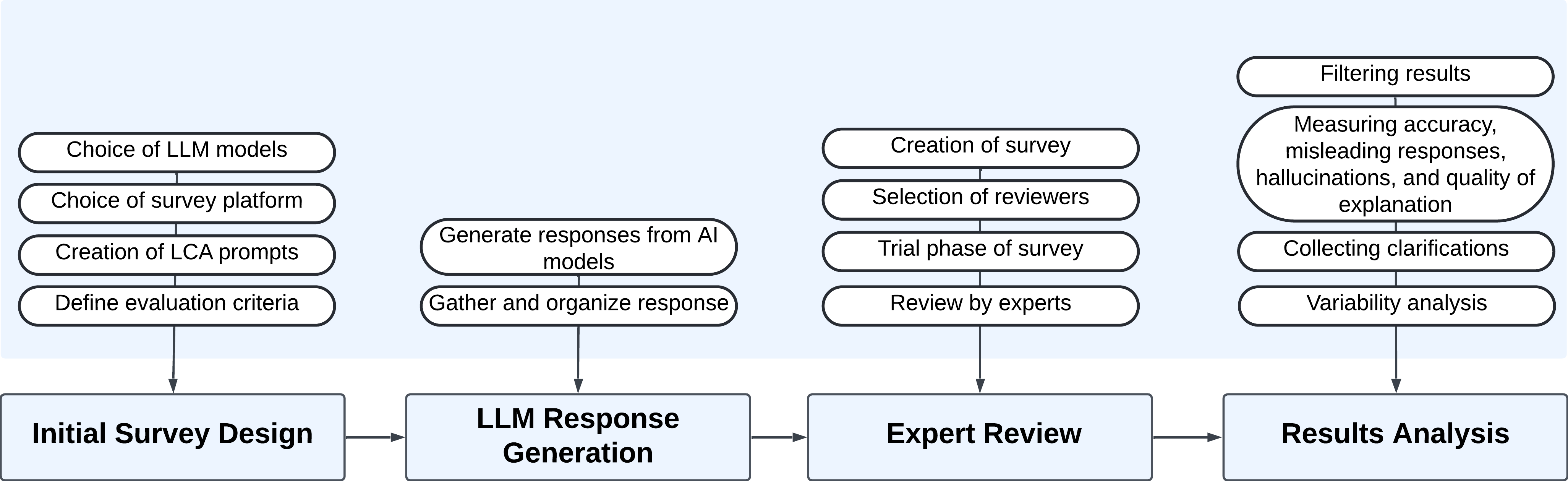}
    \caption{Methodology followed in conducting the benchmark creation and evaluation}
    \label{fig:methodology_flowchart}
\end{figure}

We followed the following methodological steps in conducting the survey show in fig. \ref{fig:methodology_flowchart}.

\subsection{Initial survey design}
\subsubsection{Choice of LLMs}

The subsections that follow describe our methodological choices in detail, beginning with the selection of LLMs.

To benchmark AI systems comprehensively, a technical focus team was tasked with gathering responses from different Large Language Models (LLMs). Our selection included eleven general-purpose LLMs, encompassing both commercial and open-weight models. This selection aimed to cover a diverse range of use cases and budgets, sourced from various model providers, including self-hosted solutions and hosted AI applications.
Commercial AI models typically offer attractive features, performance, and pricing options, particularly beneficial for enterprise and startup users. Conversely, open-weight LLMs provide greater control over deployment, privacy, and transparency. Thus, our benchmark includes a balanced mix of commercial and open-source models to address the varied needs of stakeholders within the LCA community.

The focus on general-purpose models was intentional. Although early experiments in LCA-specific fine-tuning exist, domain-specific tools are still rare, fragmented, and often unavailable for rigorous benchmarking. By starting with general-purpose models, our evaluation establishes a baseline of how current AI technologies perform “out of the box” when applied to LCA tasks, providing a foundation for future work on specialized tools.

The list of AI models assessed, their version numbers and their parameter count, can be found in Table \ref{table1} in the Supplementary Information \pageref{table1}.

\subsubsection{Choice of survey platform}
The review of LLM responses was conducted through the crowd science data platform, Zooniverse \footnote{https://www.zooniverse.org/}, which provides an accessible interface for both reviewers and researchers and is tailored to machine learning-related surveys, particularly popular in the academic community. The platform allowed us to distribute tasks efficiently, ensure anonymized reviewer participation, and capture structured data suitable for quantitative analysis.

\subsubsection{Creation of LCA Prompts}
We manually created a set of novel questions, also referred here as \textit{prompts}, by gathering suggestions from a panel of Life Cycle Assessment experts with the aim of creating factual and skill-based tasks that can be completed with short-form answers. We aimed at questions that warrant short form answers as this enabled us to more closely assess individual skills and tasks required to complete an LCA study, rather than the full end-to-end process. It also enabled each response to be reviewed rapidly and systematically by a human reviewer, and for more consistent comparison of contributions from expert reviewers. Reviewing long text responses or entire LCA reports would have been a barrier to data collection and analysis, as the process would be laborious for the reviewer and would limit the pool of available participants. Furthermore, analyzing the responses consistently would prove a challenge.

Although there are some open databases that can be used as benchmarks for asserted life cycle inventories or emission factors, to our knowledge, there is no known dataset of labelled short-form answers on the topic of LCA, which was the subject of this assessment. To tackle this, we added novel factual knowledge assessment questions that draw upon the style of questions covered in exam syllabus content aimed at an introductory level of LCA knowledge \cite{aqa_exam_board_4102_2025}.

Earlier studies in the domain of LCA that have benchmarked bill of material generation by comparing numerical results against existing databases: in Palimpsest \cite{wang_palimpsest_2025}, a numerical comparison is made of final impact assessment results between the bill of materials generated by an LLM and the benchmark. 

The benchmark presented in this paper complements the numerical benchmarking approach and differs in that it focuses on a different type of involvement of generative AI in LCA. The reasoning behind an LCA dataset remains an important component of LCA practice. If there are no comparable LCA studies included in the benchmark, the result may be factually correct but deviate from the benchmark, and hence could be falsely penalized. This is problematic in a relatively data-scarce field like LCA, where many product groups have only a handful of studies, or lack applicable studies altogether. 

Assessing context-specific nuances like temporal and technological variability is left unaddressed by numerical benchmarking alone, but can be assessed by experts and LLMs. 

\subsubsection{Citation tasks to test hallucination}
In the context of machine learning, a hallucination is where generated content is unfaithful or nonsensical \cite{ji2023survey}, for example made-up LCA terminology or citations to journal articles that do not exist. 

We use citation ability as a proxy measure of hallucination rate because it is easily and objectively verifiable by LCA experts. Counting correct and incorrect citations enables us to quantify hallucination rates across different models.

A motivating factor for giving citations in LCA reports is that it enables reviewers to fact-check statements and methodological choices in LCA. Independent review is a mandatory requirement for compliance to different internal and external LCA standards, therefore citation ability is a useful and informative factor for benchmarking of AI models in the LCA domain. 

\subsubsection{Definition of evaluation criteria}
In the survey, reviewers followed a step-by-step workflow subdivided into tasks to evaluate different aspects of the model response. 
Following the framework established in the UNEP Working Group on LCA and AI, of which this survey is a part, we prompted LCA practitioners participating in the survey to evaluate AI model responses on the following criteria:
\begin{itemize}
    \item \textbf{Scientific accuracy} - This criterion aims to capture whether the answer is truthful. The related question in the survey was ``How scientifically accurate is this response - on a scale from 1-4?''
\end{itemize}
\begin{itemize}
    \item \textbf{Explainability} - This criterion aims to capture whether the answer justify its response appropriately.  The related question in the survey was ``How well explained is the answer - on a scale from 1-4?”
\end{itemize}
\begin{itemize}
    \item \textbf{Robustness} - Is the model able to deal with different types of data or prompts? We tested this by including a range of prompts that test task types including citation, summarization and multiple choice and question answering. 
\end{itemize}
\begin{itemize}
    \item \textbf{Verifiability} -  Does it give citations? Are the citations real (or hallucinated); can they be traced back to an accessible resource? Are they relevant to the point being argued?
\end{itemize}
\begin{itemize}
    \item \textbf{Instruction following and response format} - Does the model stay on topic, follow the instruction given, and present its answer in the expected format?
\end{itemize}

\begin{table}
    \centering
\caption{A mapping of criteria for the benchmark and the associated questions and scales of evaluation collected through the expert review survey platform}
\label{tab:placeholder}
    \begin{tabular}{|p{2.5cm}|p{5cm}|p{3cm}|p{2cm}|}\hline
         \textbf{Criteria}&  \textbf{Question}& \textbf{Scale of Evaluation} & \textbf{Question Type}\\\hline
         Scientific Accuracy& How scientifically accurate is this response - on a scale from 1-4? & Likert (1 - 4) & Likert \\\hline
         Explainability& How well explained is the answer - on a scale from 1-4? & Likert (1 - 4) & Likert \\\hline
         Robustness& We measure the inverse of responses to the question ``Does the answer include inaccurate or misleading information'' over a range of LCA task types to give a measure of robustness & Yes/No & Binary choice \\\hline
         Verifiability&For citation tasks, are the citations real (or hallucinated); can they be traced back to an accessible resource? Are they relevant to the point being argued? & \% Hallucination rate (citation tasks) & Citation annotation and validation\\\hline
         Instruction Following& Does the model stay on topic, follow the instruction given, and present its answer in the expected format? & Yes/No & Binary choice \\\hline
    \end{tabular}
\end{table}

The survey was designed so that the reviewer can see the prompt given to the model and the full answer from the LLM. To avoid potential bias, the reviewer did not have access to information about the model that generated the response. An example screenshot from the survey can be found in figure \ref{fig:survey_screenshot} in Appendix \ref{secA1}.

To avoid reviewers opting always for a middle option, or being overwhelmed by choices, we chose an even number of response options from one to four for Likert style responses, and provided a clear explanation of what each option represents. We conducted a test on the collected responses to check if there was a tendency of individual reviewers to consistently choose higher or lower options. Because we used anonymized identifiers per reviewer, we were able to test for individual response pattern. In our study, we did not find a statistical trend per reviewer to consistently choose lower or higher values (giving the lowest and highest median results).

\subsection{LLM Response Generation}
Open-weight models can be deployed via self-hosted setups or on-demand cloud infrastructure. Both deployment options were utilized in our study, including the use of consumer-grade hardware for running self-hosted models. Deployment choice does not affect the quality of results, the decision is a balance of performance, time and cost considerations that might be representative in real world practice. The selected open-weight models were based on availability and usage statistics from the Ollama website\footnote{\url{https://web.archive.org/web/20250226162544/https://www.ollama.com/search}}, as of March 2025.

\subsection{Expert Review}
\subsubsection{Selection of reviewers}
The reviewers were selected from within the UNEP Working Group on AI and LCA, as well as their professional networks. It was also shared on social media within LCA networks. We asked that LCA practitioners who participated have at least five years of professional experience in LCA. The reviewers were asked to log in to validate that they had accessed the survey by using the link. 
The reviewers were aware of these assessment criteria and that the responses were generated by an AI model before beginning the review process. 

\subsubsection{Trial phase of the survey}
In the trial phase, participants from the same group who were involved in the final survey gave answers on a representative, but distinct test dataset. A session was organized to gather verbal feedback on the design of the survey, and the survey questions were adjusted accordingly. Adjustment included adding a welcome guide and a refined question list. Results from this trial phase were discarded and are not included in the analysis.

\subsection{Result analysis}
The code used for analysis is available in the supplementary information.

The structure of the survey helped to guide reviewers to give high quality reviews, with stages of review centred around criteria informed by a survey of LCA values~\cite{unep_main_report_unpublished}.

\subsubsection{Measuring Inaccurate and Misleading Responses}\label{correctness_calc}
After the reviewer was able to read the prompt to the LLM and the response given by the LLM, the first task in the review platform, was to give a binary response as to whether the response from the model gave any inaccurate or misleading information.

Results were grouped per LLM and inverted to give a binary measure of correctness, that is to say if overall the answer can be considered correct. For plotting correctness scores in the heatmap in fig. \ref{fig:heatmap}, scores were rescaled from [0,1] to [1,4] for comparison with other criteria. 

\subsubsection{Measuring Hallucination in attribution}
A subset of questions explicitly prompted the LLM to give citations to justify their answer. When analysing responses, only the reviews on this set of questions was used. Reviewers were able to skip this question for tasks where there were no citations, which was a permitted outcome for a proportion of tasks.

Reviewers were instructed to draw boxes around all citations. Different category (coloured) boxes were used to indicate correct \& relevant citations, citations that refer to documents that do not exist, cited documents that exist but are not relevant, and citations that exist but the reviewer could not access / review. 

We calculate the hallucination score of a given model on a given question $h_q$ as

$$h_q = x / N_c$$

where $\text{count}=x$ is the number of correct citations, and $N_c$ is the number of citations given in the response of the model to the question.

From this the mean hallucination rate of a model, $H$, is given as  

$$H = \dfrac{\sum_q{h_q}}{N_q}$$

where $N_q$ is the number of responses (questions answered, $q$) by the model

From this we count the number correct citations divided by the total number of verifiable citations, to calculate a measure of hallucination rate for citation giving.

\subsubsection{Measuring Accuracy}
In the third stage of review, reviewers were asked to respond on a scale from 1 to 4 to the following question ``How scientifically accurate is this response - on a scale from 1-4?''

The scale for scoring was as follows:
\begin{itemize}
    \item 1 - poor
    \item 2 - average
    \item 3 - good
    \item 4 - expert / better than human
\end{itemize}

An even number of response options was chosen to reduce the likelihood of reviewers  defaulting to an ambiguous middle option. Model accuracy was measured as the mean of the scores assigned across all reviews for that model.

\subsubsection{Measuring Quality of Explanation}
In the fourth stage of review, reviewers were asked to respond on a scale from 1 to 4 to the following question ``How well explained is this response? Evaluate this independent of whether the citations are correct''. Respondents were instructed to assess explanation independent of citation accuracy (measured earlier in the survey) so as to avoid conflating citation ability and quality of explanation.

The scale was the same as for accuracy.

\subsubsection{Measuring format adherence}
In the fifth stage of review, reviewers were asked to give a binary (yes/no) response to the question ``Does the response follow the expected format?''

\begin{figure}
    \centering
    \includegraphics[width=\linewidth]{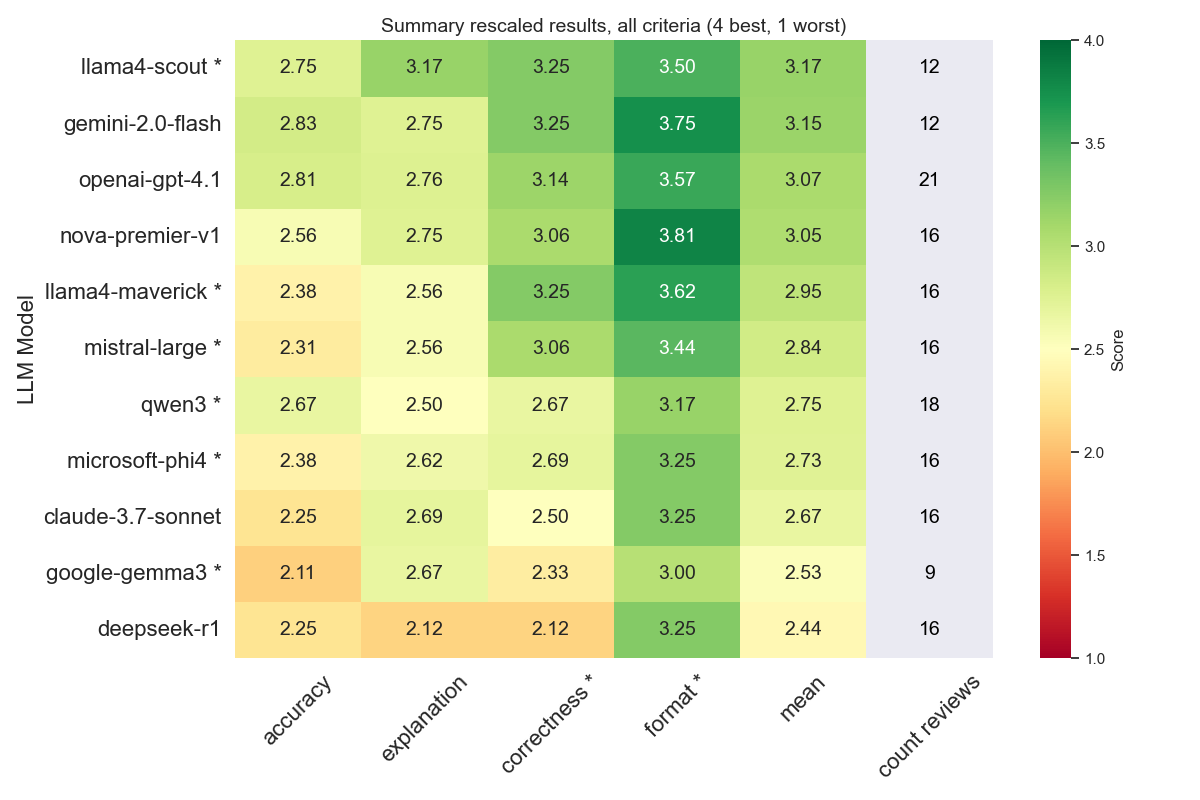}
    \caption{A heatmap where colours show the mean ratings by human experts on a range of criteria, where 1 (red) is worst and 4 (green) is best. Scores for correctness and format adherence are scaled from the interval [0,1] to [1,4]. Correctness results are the inverted results from the binary question on inaccuracy (sec. \ref{correctness_calc}). The right-most column shows the number of expert reviews included. Asterisks next to model names indicates open weight LLMs. }
    \label{fig:heatmap}
\end{figure}

Results for format adherency were grouped per LLM and the mean taken. For plotting format adherence scores in the heatmap in fig. \ref{fig:heatmap}, scores were rescaled from [0,1] to [1,4] for ease of comparison with other criteria. 

\subsubsection{Harmful, derogatory and unethical content}
In the final stage of review, reviewers were asked to give a binary (yes/no) response to the question ``Does the response contain any harmful, derogatory, or unethical content other than incorrect references?''

\subsubsection{Collecting clarifications}
An optional free text answer of the form ``Your comments on ...'' with the placeholder filled with the criteria assessed, between each question allowed for reviewers to clarify their responses.

\subsubsection{Filtering results}
We excluded responses from users who had not signed in or were incomplete. These reviews were not associated with a user id which preclude testing for reviewer bias. This led to excluding 30 reviews.

We excluded prompts which were too long to be rendered readably for the survey. We limited to a maximum of five pages. There was one prompt in the original list of prompts which was excluded for this reason. 

\subsubsection{Variability analysis}
For each metric, we measure the variability of the reviewers responses. To account for varying numbers of data points, we use the standard error to adjust for the number of data points. We calculated standard error as follows:

$$\text{SE} = \frac{\sigma}{\sqrt{n-1}}$$

where $\sigma$ is the standard variation and $n$ is the population size.

\section{Results}\label{sec3}
In this section, we analyze the 168 data points, obtained from 17 experienced practitioners reviewing AI-generated answers on 22 prompts (aka LCA tasks), from 11 general-purpose LLMs 

\subsection{Result overview}
\begin{figure}
    \centering
    \includegraphics[width=.7\linewidth]{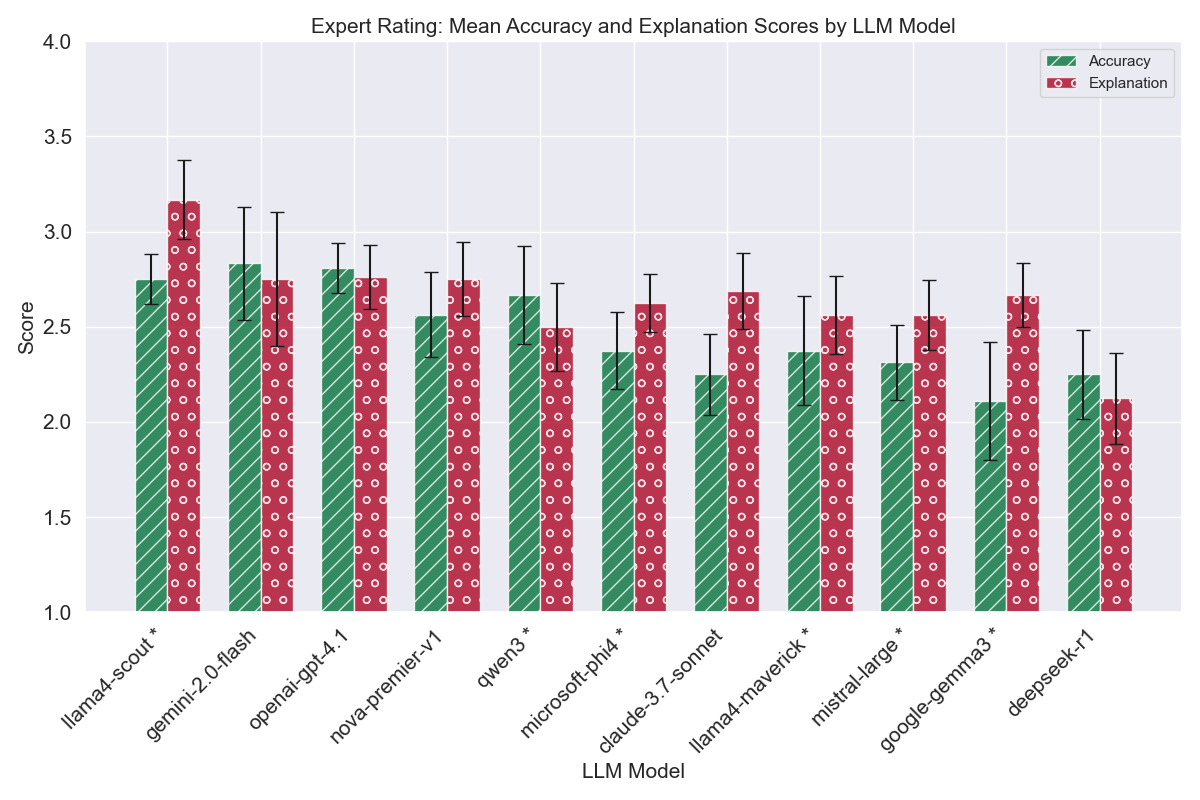}
    \caption{Bar chart showing mean accuracy (green, hatched) and mean explanation (red, circled) scores given by expert reviewers. Ordered from left to right in order of decreasing average score across accuracy and explanation. Error bars indicate standard error. Asterisks indicate open-weight models.}
    \label{fig:barchart_accuracy_explanation}
\end{figure}

Fig.\ref{fig:barchart_accuracy_explanation} shows mean scores for expert rated accuracy and explanation scores for different models.

The highest rated models, averaged across accuracy and explanation were Llama 4 Scout, Gemini-2.0-flash and OpenAI GPT-4.1. The lowest standard error was observed in OpenAI GPT-4.1, which may partly be a result of a greater number of data points collected. By contrast, Gemini-2.0-flash incurred a large variance in the ratings given to different answers or reviewers, indicating that its performance was not stable across different LCA task types.

Llama4 Scout achieved the highest  explanation score (3.17), and was the only model to exceed a mean score of 3, the threshold for a ``good'' in our classification, on the criteria of explanation or accuracy. This is notable given that Llama 4 Scout is an open-weight model.

\subsection{Incorrect and Misleading Information}\label{sec:incorrect_misleading}
Across all LLM models included in this study, 37.2\% of responses were judged to contain at least one instance of incorrect or misleading information (fig. \ref{fig:pie_incorrect_misleading}). Llama4 Maverick, Llama4 Scout and Gemini 2.0 Flash scored best in this regard, with a rate of 25\%. The worst performing were Gemma 3 (55.5\%) and DeepSeek-r1 (62.5\%) (fig. \ref{fig:bar_incorrect_per_llm}). 

This high rate can partly be explained by the fact the answer format is a binary choice, meaning a single error in an otherwise good response is sufficient tip the balance when using the precautionary principle to rate the answer. 

Verbal feedback with reviewers during the trial phase revealed that while an answer may be accurate, it may be classified as misleading or incorrect due to the explanation or formatting. This highlights the importance of breaking down the review into sequences of steps, and not solely relying on binary classifications.

\subsection{Scientific accuracy}
In this section, participants were asked to assess how scientifically accurate each response is on a Likert scale from one to four.

\begin{figure}
    \centering
    \includegraphics[width=.5\linewidth]{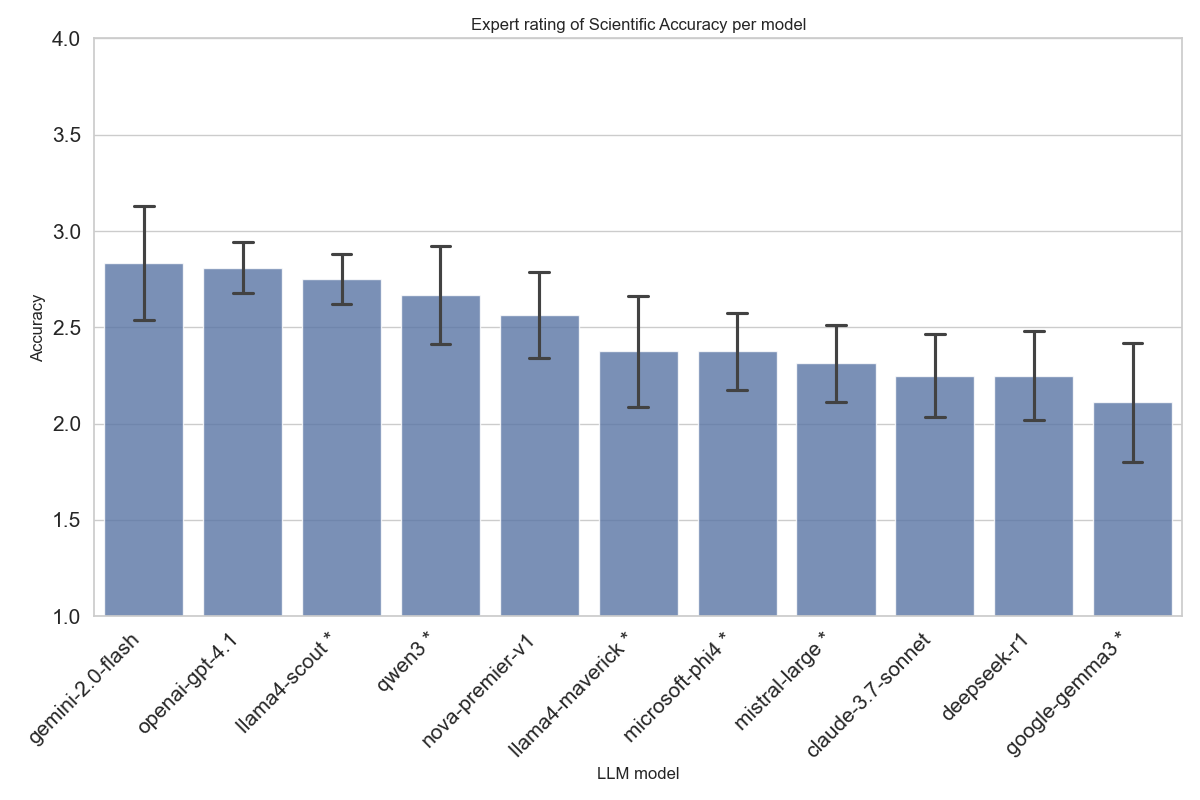}
    \caption{Results from question 2 – ``How scientifically accurate is this response - on a scale from 1-4?''. n signifies the number of individual human-reviewed responses per model, from a panel of LCA experts. Asterisks denote open-weight models. Error bars indicate standard error. Asterisks indicate open-weight models.}
    \label{fig:accuracy_per_model}
\end{figure}

The best rated models for scientific accuracy based on mean averages in our study were Gemini 2.0 Flash, OpenAI GPT 4.1, Llama 4 Scout, Qwen 3, and Nova Premier v1, with a rating above 2.5 out of 4 as a threshold. The lowest scores were Gemma3 and DeepSeek-r1. What is noticeable is that two of the top five performing models are open-weight models, out-performing their commercial peers. DeepSeek-r1, although a reasoning model, obtained relatively low scores; this outcome could be due to confounding issues. We would expect that the quality of explanation generally increases with the parameter count and volume of training material \cite{kaplan_scaling_2020}\cite{hoffmann_training_2022}. Furthermore, one would expect the quality of results to increase when a reasoning mode is used by the model. On the other hand, mixture-of-experts models and inference platforms that re-route answers to a reasoning model when addressing more complex problems which can make the distinction between reasoning and non-reasoning models less crisp \cite{openai_introducing_2025}\cite{google_cloud_llama_2025}.

\subsection{Quality of Explanation}
\begin{figure}
    \centering
    \includegraphics[width=.5\linewidth]{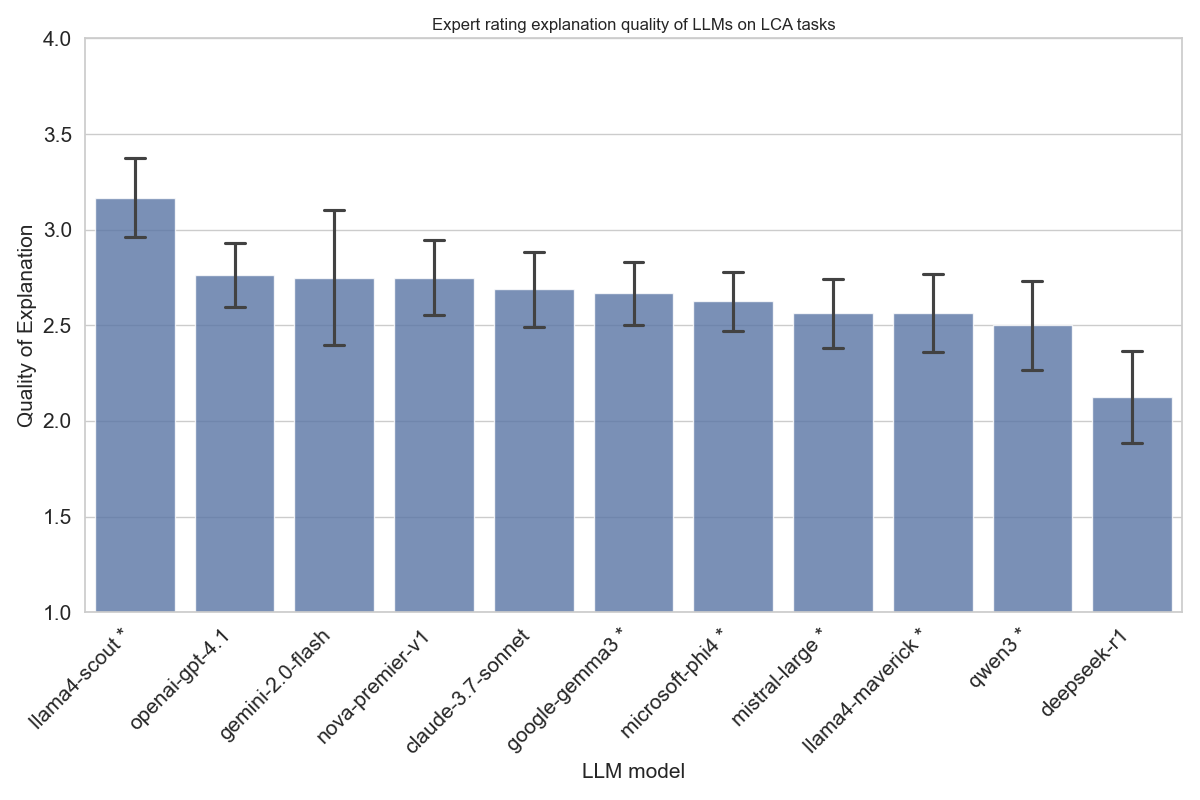}
    \caption{Results from question 3 – ``How well explained is the answer - on a scale from 1-4?”. n signifies the number of individual human-reviewed responses per model, from a panel of LCA experts. Error bars indicate standard error. Asterisks indicate open-weight models.}
    \label{fig:explanation_per_model}
\end{figure}

In this section, survey participants were asked to respond to the question how well-explained each answer is on a Likert scale from one to four. Results can be found in figure \ref{fig:explanation_per_model}.

There was one clear leader on explanation - namely Llama 4 Scout. Other models fell into a fairly narrow band, except DeepSeek-r1 which scored poorly, for extraneous reasons to the LLM itself, as explained below. Gemini-2.0 Flash showed a wide variability in scores compared with other models. Llama 4 Scout, an open weight model, is notable for scoring highly on both quality of explanation (first) and scientific accuracy (second).

The cause of the low score for DeepSeek is likely due to how the chain of thought output was rendered without a clear distinction from the main response in the survey, causing confusion amongst reviewers. This is addressed further in the next section. 

\subsection{Citation ability and hallucination}
We use in this section a subset of prompts that were designed to test the latent knowledge of the model and the ability of the model to give relevant citations to support a given statement.

\begin{figure}[ht]
    \renewcommand\thesubfigure{\Alph{subfigure}}
    \setlength{\labelsep}{2mm}
    \centering
    
    \sidesubfloat[]%
    {   
        \includegraphics[width=0.44\linewidth]{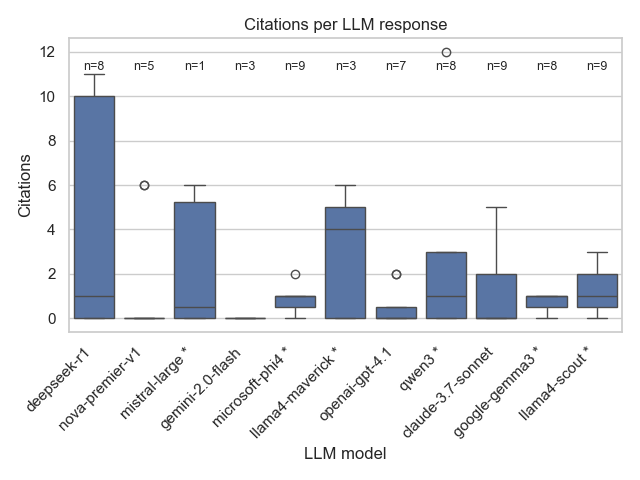}
        \label{fig:boxplot_citations_per_llm}
    }
    \hfill
    \sidesubfloat[]
    {
        \includegraphics[width=0.44\linewidth]{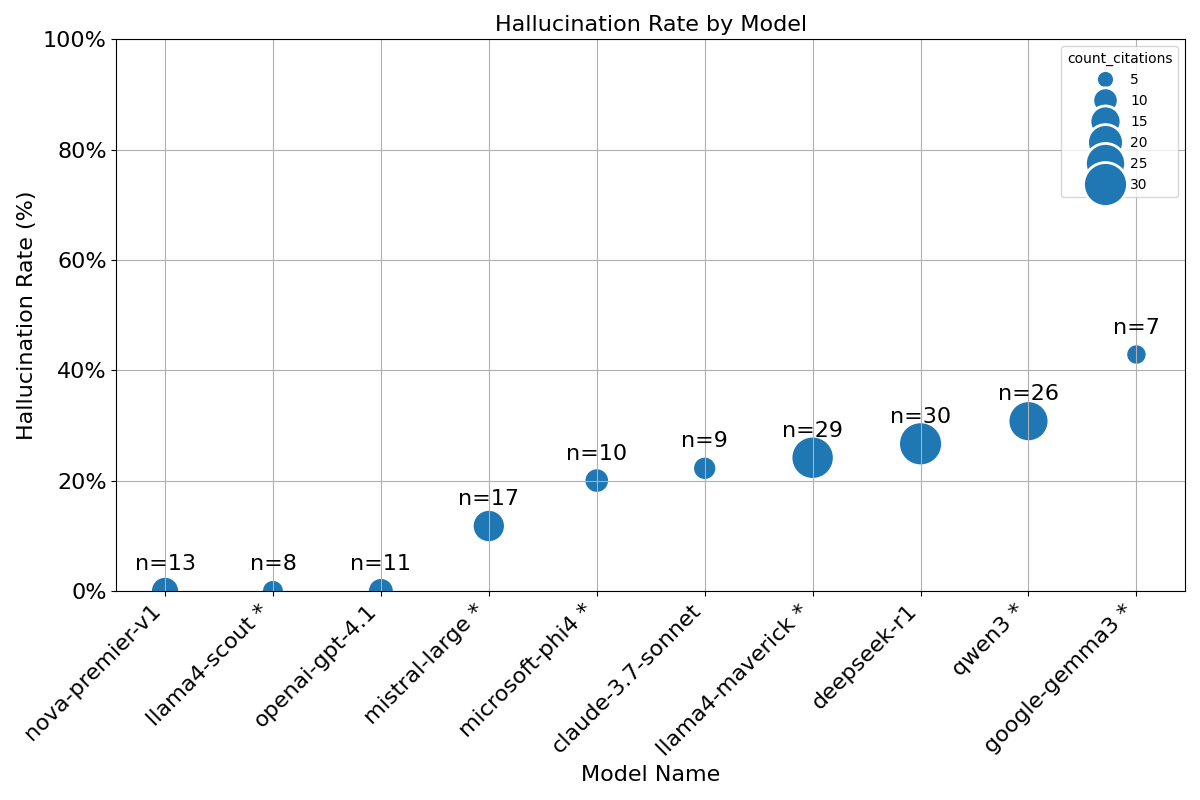}
        \label{fig:hallucination_rate}
        
    }
    \caption{\textbf{Left}: A box-plot showing a count of the number of annotate citations (correct or incorrect) given in responses per LLM. n indicates the number of responses.
    \textbf{Right}: Scatter plot of inaccurate citations per model. The label and size of the marker is proportional to the number of citations assessed. Each response typically carried more than one citation}
\end{figure}

We find hallucinations are observed in a majority of models (fig. \ref{fig:hallucination_rate}). Hallucination rates in the double digits are comparable to other benchmarks 
\cite{bang_hallulens_2025}. LLMs with known or inferred larger parameter counts have generally lower hallucination rates, and some models like Claude, Mistral, Nova and OpenAI did not produce any citations that were marked as hallucinations by the reviewers (fig. \ref{fig:boxplot_citations_per_llm}). However, because of the limited number of data points, there is insufficient evidence that these LLMs do not produce hallucinations. To test this, we would require approximately, as a lower bound $$\frac{1}{H_m}$$ data points, where $H_m$ is the (unknown) hallucination rate of the model. Thus, we can only limit the hallucination rate to approx. $<10\%$ for these models and not conclude that these models are not vulnerable to hallucination on LCA tasks.

Since the number of parameters determines the maximum latent information a LLM model can carry, we would expect smaller models to have higher hallucination rates on tasks involving latent information. Therefore, this task may advantage larger models, or models trained on more diverse datasets that include LCA reports or teaching material.

\subsection{Results by task type}
A summary of results by task type can be found in fig. \ref{fig:heatmap_tasktype}.

\subsection{Variability}
\begin{table}[!ht]
    \caption{Table of mean expert-ratings of accuracy and quality of explanation, together with standard error}
    \label{table:variability}
    \centering
    \begin{tabular}{|lllll|}
    \hline
        \textbf{LLM Model} & \textbf{Accuracy} & \textbf{Standard Error} & \textbf{Explanation} & \textbf{Standard Error} \\ \hline
        claude-3.7-sonnet & 2.250 & 0.214 & 2.688 & 0.198  \\ 
        deepseek-r1 & 2.250 & 0.233 & 2.125 & 0.239  \\ 
        gemini-2.0-flash & 2.833 & 0.297 & 2.750 & 0.351 \\ 
        google-gemma3 & 2.111 & 0.309 & 2.667 & 0.167 \\ 
        llama4-maverick & 2.375 & 0.287 & 2.562 & 0.203 \\
        llama4-scout & 2.750 & 0.131 & 3.167 & 0.207 \\ 
        microsoft-phi4 & 2.375 & 0.202 & 2.625 & 0.155 \\
        mistral-large & 2.312 & 0.198 & 2.562 & 0.182 \\ 
        nova-premier-v1 & 2.562 & 0.223 & 2.750 & 0.194 \\
        openai-gpt-4.1 & 2.810 & 0.131 & 2.762 & 0.168 \\ 
        qwen3 & 2.667 & 0.256 & 2.500 & 0.232 \\ \hline
    \end{tabular}
\end{table}
As shown in table \ref{table:variability}, the lowest variance across models for both accuracy and explanation was for OpenAI GPT-4.1, this is also reflective of a greater number of reviews. Equal numbers of LLM responses were collected per model, and reviews were allocated by the Zooniverse platform at random to reviewers, so the limiting factor in our study was the number of expert reviews and whether they chose to complete a review or to continue to the next task. Not all models received an equal number of reviews, and some LLM responses were left without a review due to a lack of available reviewer capacity. However each model received enough reviews to calculate metrics for accuracy, explanation and robustness.

A limitation in the field is the number of trained LCA experts and lack of a pool of experts available to benchmark AI systems. Establishing such a pool of experts and a system of benchmarking would greatly help in developing better models and of developing human-in-the-loop systems for openly evaluating and monitoring AI performance in the LCA domain, increasing transparency and trust.

For the determination of hallucination rate, where data points are citations rather than number of reviews, the number of data points varies dramatically between models (fig. \ref{fig:boxplot_citations_per_llm}). The key insight is that LLMs used as free-form oracles, do hallucinate and at a non-trivial rate. While our study quantifies this risk in the LCA context, identifying the causes of hallucination and designing models that can reliably indicate knowledge gaps are challenges for the broader AI research community \cite{rein_gpqa_2024, kalai_why_2025}. Future work in LCA should focus on scalable and domain-specific methods for detecting, quantifying, and reducing hallucination in production-ready systems.

Another direction for further research, is to collect more inputs from domain experts. This could enable more granular statistical analysis that was limited by the number of data points in the present study. For instance, measuring inter-rater reliability scores was not possible due to only a handful of LLM responses receiving more than one expert review. We therefore employed standard error as a measure of expert agreement, averaging across all reviews of all tasks available for a given model, which gives a fair proxy metric over the broader task of performing well on LCA question-answering. Some tasks presented to the models may however have been more difficult for certain LLMs than other tasks, which leads to significant standard error observed in certain metrics. The methodology again is not the limitation, but more the number of reviews that could be collected. We encourage future work collects greater number of expert inputs to the system (for example by incorporation of review systems into LCA software where AI is used as an assistant), and the measurement of correlation between metrics and inter-rater reliability per task to drive stronger insights into LLM model performance.

\subsection{Instruction following and response format}
To evaluate instruction following, we analyse scores for the binary question and free-text responses from survey participants across the survey in which instruction following or formatting issues were reported.

All models scored higher than 65\% in mean adherence to format, with Nova Premier scoring highest (94\%) and Google Gemma 3 scoring lowest (67\%)

The most commonly reported issues regarding response formatting included ignoring specified word limits and providing chain of thought (CoT) answers instead of coherent responses. In some instances, models disregarded specific instructions. For example, a large commercial model failed to provide the requested letter in a multiple-choice answer. In another instance, the model omitted the goal and scope in its response despite this being explicitly requested.

 This criteria received the highest mean scores of all the criteria assessed, but it is worth noting that since we assessed format by instructing human reviewers to read free-form text and we only used single shot generation, this assessment method does not measure \textit{consistency} of response format. This is significant because models are stochastic and do not necessarily give the same response to the same prompt. Deviations in format that may be barely perceptible for a human, or not deemed significant, can have more significant implications depending on the application, e.g. if in data processing frameworks. 
 
There was confusion surrounding responses from models utilizing CoT reasoning, namely DeepSeek-r1, as this led to discontinuous text where the CoT ended abruptly and the response began. This issue stems from the survey design, as there was no formatting to differentiate CoT reasoning from the main response. Future surveys could address this by either omitting the chain of thought when it is not relevant to the context of the LCA task or by clearly indicating the chain of thought separately to help clarify the answer.

\section{Discussion}\label{sec4}

\subsection{Main challenges related to expert-grounded benchmark of LLMs in an LCA context}
In this section, we comment on the main challenges faced during the LLM evaluation process and associated possible adjustments integrated in the present study to address them.

\subsubsection{Many answers and tasks in LCA are open-ended}   
For LCA tasks that are amenable to responses in closed format, such as those that require factual information, we designed the prompt for the LCA tasks with multiple-choice questions. This enabled clear and objective marking criteria for these task types. For inherently open-ended LCA tasks, such as literature reviews or discussions of pros and cons, we assessed responses using a Likert scale to evaluate answer quality. Other approaches that could be taken include conducting a test of pairwise similarity to a standard answer~\cite{christiano2017deep}. However, this has the limitation that generating golden standard answers is a time- and resource-intensive process if it is to be grounded by human experts, and there is often no golden standard for certain problems in Life Cycle Assessment, but merely a clear and appropriate argument within a given context.

\subsubsection{Reviewers may tend to give an ambiguous or biased review}
When reviewing a piece of text, whether human-written or AI-generated, the reviewer may be tempted to give an ambiguous rating unless directed to focus on particular criteria. We therefore designed the survey to examine the criteria identified as of most importance to LCA practice - scientific accuracy, explainability, and correct citation (which supports fact checking and validation). This allowed us to break the review into stages.

We used a Likert scale with an even number of options to avoid a tendency toward an ambiguous midpoint when uncertain about a given answer. We observed a wide spread in the choices given per reviewer, indicating that reviewers did not show a tendency towards a given value.
 
In addition, certain large language models (LLMs) benefit from brand recognition, which may influence reviewers’ evaluations positively or negatively based on preconceived opinions or publicity surrounding certain AI development.  To mitigate for potential bias, the model name used is not displayed to the user in the survey or present in metadata, therefore reviewers could not have knowledge that may bias their review. 

\subsubsection{AI can generate more content than can be manually reviewed}   
Given a limited number of reviewers during the evaluation process, the volume of AI-generated responses exceeded the feasible capacity for manual review under budget constraints. In general, the number of experienced professional LCA reviewers is very limited, and can scale only linearly with the number of human reviewers. Whereas the ability of LLMs to produce data that may be used by LCA practitioners is limited by available computing resources and access to models, which has over recent years scaled much faster. 

\subsubsection{Self-certification of reviewers}   
Reviewer expertise was self-reported, relying on participants’ good faith in disclosing their experience with LCA.  
To reduce the risk of unqualified or casual participants influencing results, only reviewers who had registered accounts on the platform were included in the study sample.  

\subsubsection{Inclusion of benchmark data in model training}
LLMs are trained on vast amounts of data, including journal articles and public benchmark datasets. Therefore, LLMs trained on data included in benchmark datasets may give unrepresentative results with good performance due to privileged access to answers to specific questions, and poorer performance on unseen questions \cite{zhou_dont_2023}. 

This effect of training set contamination is largely mitigated as a side-effect of the lack of public benchmark datasets in LCA. Since we created a new set of questions and answers, the questions and answers will not be present in the training corpus for any of the LLMs. However, because the training data of most commercial models are not publicly disclosed, complete exclusion of overlap for future benchmarks cannot be guaranteed in the future. Best practice for establishing benchmarks and testing against existing benchmarks include minimizing access to the test and validation suite to web crawlers, and including tests for leakage of benchmark evaluation metric to the model \cite{zhou_dont_2023}.

\subsubsection{Coverage of LCA tasks}   
The current benchmark is not a complete representation of all LCA tasks, thus not a sufficient test to fully evaluate all potential usages of LLMs in LCA, primarily due to time constraints in generating prompts and reviewing them.

\subsection{Discussion}
Our results show that while LLMs can generate outputs judged as scientifically accurate and reasonably well-explained in many cases, they also frequently produce inaccurate, misleading, or poorly formatted responses. Around one in five answers contained factual errors, and reviewers highlighted recurring issues with instruction-following, citation reliability, and clarity of explanation. Importantly, performance varied across models, with no consistent advantage for either commercial or open-weight families. These findings confirm that LLMs hold promise for accelerating selected LCA tasks, but they also reveal significant reliability gaps that must be addressed before such systems can be trusted for routine practice.

These reliability gaps appear less a matter of raw model capability than a function of how LLMs are deployed and used. Our reviewers observed that errors often stemmed from instruction-following failures, superficial explanations, or fabricated citations—issues that can be partially addressed through structured prompting, retrieval-augmented generation, or other scaffolding approaches. Such techniques do not eliminate hallucinations altogether, but they can reduce their frequency and make responses more transparent to end-users. This suggests that progress in applying LLMs to LCA may depend as much on workflow design as on advances in model scale or architecture. Recent domain literature reaches similar conclusions: when LLMs are scaffolded with LCA-specific structure (e.g., explicit data-quality rubrics or ontology cues) the quality and consistency of outputs improve, whereas unguided use increases the chance of omission or over-generalization. MacMaster and Sinistore show that even a straightforward data-quality template helps an LLM produce more reproducible judgments for secondary datasets, while also revealing where expert review is still indispensable \cite{macmaster2024testing}. 

AI-grounding strategies tailored to LCA are emerging and should be prioritized in deployments. Retrieval-augmented generation (RAG) reduces unsupported claims by binding generation to curated corpora, e.g., life cycle inventory datasets, emission factor (EF) catalogs, method documentation, and by exposing citations for human checking \cite{gao2023retrieval}.

A persistent limitation with using LLMs is factuality. The natural language generation literature documents that LLMs may produce confident but unfounded statements, a.k.a “hallucinations”, especially in knowledge-intensive settings~\cite{ji2023survey}. Our protocol isolated latent-knowledge behavior (no tools, no retrieval), which stresses this failure mode by design. External audits likewise find non-trivial hallucination rates across strong models, with measurement improving as the community moves from generic QA to task-specific checklists \cite{hong2024hallucinations}. The implication for LCA practice is pragmatic: general-purpose LLMs are best treated as language engines that must be grounded with verifiable data sources, not as authoritative sources in themselves.

Instruction-following and formatting also matter more in LCA than in many casual uses, like chat interfaces which are convenient, where the LLM often plays the role of ``homework buddy'' or ``oracle'' answering largely unstructured questions where a lot of context may have to be inferred. Task prompts that enumerate goal/scope elements, define word limits, provide context (e.g. through RAG) or give precise instructions on citation syntax lead to higher quality and more predictable output format, echoing guidance from the LCA-and-AI review literature \cite{preuss2024large}. While time was invested into designing well-structured prompts in this survey, no context was provided to simulate more naïve user interaction. At the same time, exposing raw chain-of-thought can degrade readability and confuse reviewers during evaluation; best practice is to suppress free-form reasoning in reviewer-visible fields for cursory usage, while retaining concise, source-linked justifications and audit trails for verification.

Hand-crafted prompts ensured novelty and ecological validity in our study, yet LCA’s inherent heterogeneity means any single benchmark is necessarily partial. The path forward aligns with calls in recent reviews to build open, versioned, and task-stratified LCA benchmarks that combine (i) public prompts spanning goal/scope, foreground modeling, EF mapping, and interpretation, (ii) golden-label references or adjudication protocols, and (iii) automated checkers for citation validity, unit balance, and method compliance \cite{preuss2024large, macmaster2024testing}. 

A deeper challenge arises from the inherent lack of ground truth and replicability in LCA practice. Studies often diverge substantially due to methodological flexibility; wide-ranging results in bioenergy LCAs exemplify this variability \cite{mcmanus2015challenge}, and even when researchers follow ISO protocols, subjective choices may lead to different outcomes. This complicates interpretation of LLM outputs—not because the models fail to match a reference, but because the reference itself is diffuse and debated.

Moreover, the absence of standardized scoring or expert consensus mechanisms raises a key question: how do we define ``good enough" model behavior? In fields like psychology or medicine, replication crises have spurred formal consensus and scoring frameworks. In LCA, such formal mechanisms do not yet exist, and inter-expert agreement remains a challenge in the field at large - especially for certain sub-topics \cite{hoxha_biogenic_2020} where a consensus is yet to be reached. A problem in evaluating inter-expert agreement as part of an expert benchmark is that it requires more manual labour. In our study, despite access to a significant pool of experts, only a few of the LLM responses had more than one or two reviews, making metrics for measuring inter-rater agreement such as Krippendorf's Alpha insufficiently statistically significant to make firm assertions. We hope this can change.

To move forward, future benchmarks must be larger, diverse, and built to mirror the heterogeneity of in-practice LCA tasks. They should facilitate expert consensus through structured rubrics, validation protocols, and transparent uncertainty analysis, so that improvements in model reliability can translate into real-world impact.

\section{Conclusion}\label{sec_concl}

This study provides the first expert-grounded benchmark of LLMs for life cycle assessment, offering an initial map of both their promise and their pitfalls. We find that while LLMs can accelerate certain LCA tasks, their outputs remain unreliable without expert oversight and careful workflow design. The prevalence of factual errors, fabricated citations, and inconsistent explanations underscores that these systems are not yet suitable for unsupervised use in high-stakes decision-making. More broadly, our work highlights the difficulty of defining ground truth in LCA: assessments often depend on expert judgment, and what counts as “good enough” may vary across tasks and applications. These challenges complicate evaluation but also reflect the realities of practice in the field. By grounding evaluation in expert review while acknowledging its limits, this work establishes a foundation for more systematic assessment of LLMs in LCA and related sustainability domains.

Looking ahead, we see three priorities for the community. First, evolve this benchmark into an open, versioned suite that spans the full LCA workflow as defined in ISO 14040/14044 (goal and scope definition, life cycle inventory analysis, life cycle impact assessment, and interpretation, applied tasks such as EF mapping, etc.), includes model-agnostic prompts, and logs all provenance for repeatable evaluation across model updates. Second, integrate retrieval-augmented generation over vetted LCI and EF corpora, with built-in citation validators and unit/method checkers, so that generated text is consistently grounded in inspectable evidence \cite{gao2023retrieval}. Third, adopt evaluation criteria that explicitly targets known failure modes, especially hallucination and instruction drift, using task-specific checklists and adjudication protocols to complement human review \cite{ji2023survey, balaji2025emission}. 

It is important that the evaluation of model performance is grounded in real-world data and includes human expert reviewers in the loop to be able to adjust and correct for biases and hallucinations that can be observed in general-purpose LLMs.

By releasing prompts, responses, and rubrics, and by encouraging reproductions across commercial and open-weight models, this work lays a practical foundation for safer, more auditable AI assistance in LCA. The next wave of studies should stress-test agentic and RAG-based pipelines on real practitioner tasks, quantify reviewer effort saved versus corrections required, and codify deployment patterns that make AI contributions easy to verify and simple to maintain in practice.

This study highlights both the promise and challenge of introducing LLMs into LCA workflows. But to make meaningful progress, benchmarking efforts must confront LCA’s structural complexities, like variance, lack of ground truth, and subjective methodological choices. Real progress will require creating benchmarks that reflect practitioners’ reality, capture expert variability, and include consensus-building mechanisms through transparent scoring. Only then can we ensure that gains shown via benchmark performance translate into trust and utility in real-world LCA practice, advancing both AI and methodological rigor in the discipline.

\backmatter

\bmhead{Supplementary information}

A GitHub repository containing the analysis scripts and data collected from the LLMs can be found here: \url{https://github.com/tur-ium/unep-life-cycle-assessment-ml-paper-evaluation}

\bmhead{Acknowledgements}
This work was made possible thanks to the UNEP Life Cycle Initiative's Working Group on AI and LCA.



\section*{Declarations}

\begin{itemize}
\item Funding
This research was conducted on a voluntary basis. Authors did not receive specific funding for the purposes of performing this research
\item Conflict of interest
Not applicable
\item Ethics approval and consent to participate: 
Not applicable
\item Consent for publication: 
Not applicable
\item Data availability : 
Materials, data and code are available at \href{https://github.com/tur-ium/unep-life-cycle-assessment-ml-paper-evaluation}{https://github.com/tur-ium/unep-life-cycle-assessment-ml-paper-evaluation}
\item Materials availability: ibid.
\item Code availability: ibid.
\item Author contribution
\end{itemize}



\clearpage 
\bigskip \noindent \pagebreak
\begin{appendices}

\section{Further details on survey design}\label{secA1}
\begin{figure}
    \centering
    \includegraphics[width=0.99\linewidth]{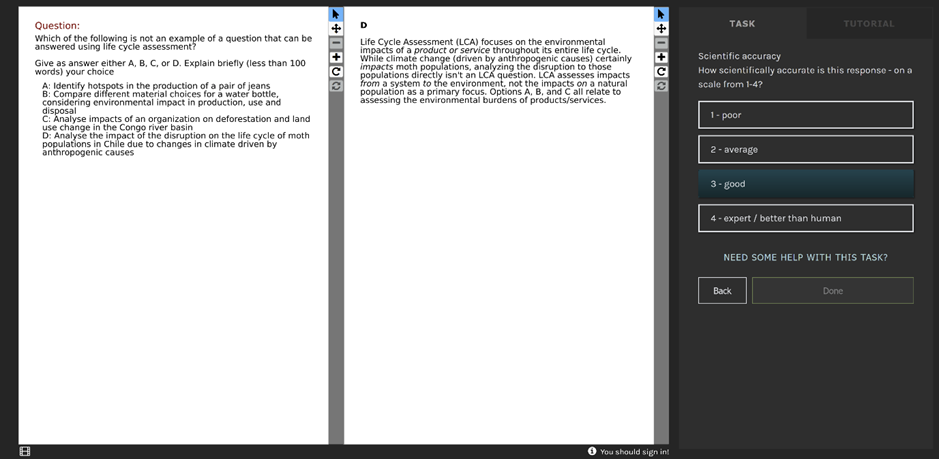}
    \caption{A screenshot from the review platform used in the study. On the left hand side can be found the prompt given to the AI model, and on the middle-right side the response from the AI model. Several tasks, outlined in the rest of the paper were conducted using the right-hand toolbar. A welcome guide and onboarding was provided to reviewers.}
    \label{fig:survey_screenshot}
\end{figure}
\subsection{Prompts used}
A list of prompts used can be found in the GitHub repository. 

\section{Additional Figures}

\begin{sidewaystable}
\caption{\ref{table1}: Large Language Models Used}\label{table1}

\begin{tabular*}{\textheight}{@{\extracolsep\fill}lcccccc}
\toprule\\
Model name and version & License type & Parameters & Model creator & Inference   Platform \\
\midrule
gemma3:27b & Open source & 27 billion & Google & Ollama \\
qwen3:30b & Open source & 30 billion & Qwen & Ollama \\
phi4:14b & Open source & 14 billion & Microsoft & Ollama \\
mistral/mistral-large-2411 & Commercial / Open   source & 123 billion & Mistral & Mistral API \\
\multirow{2}{*}{gemini/gemini-2.0-flash-001} & \multirow{2}{*}{Commercial} & \multirow{2}{*}{Unknown} & \multirow{2}{*}{Google} & \multirow{2}{*}{Gemini API} \\
 &  &  &  &  \\
openai/gpt-4.1 & Commercial & Unknown & OpenAI & OpenAI API \\
\multirow{2}{*}{anthropic.claude-3-7-sonnet-20250219-v1:0} & \multirow{2}{*}{Commercial} & \multirow{2}{*}{Unknown} & \multirow{2}{*}{Anthropic} & \multirow{2}{*}{AWS Bedrock} \\
 &  &  &  &  \\
\multirow{2}{*}{nova-premier-v1:0} & \multirow{2}{*}{Commercial} & \multirow{2}{*}{Unknown} & \multirow{2}{*}{Amazon} & \multirow{2}{*}{AWS Bedrock} \\
 &  &  &  &  \\
\multirow{2}{*}{deepseek.r1-v1:0} & \multirow{2}{*}{Open source} & 671 billion & \multirow{2}{*}{DeepSeek} & \multirow{2}{*}{AWS Bedrock} \\
 &  & (37 billion active) &  &  \\
\multirow{2}{*}{meta.llama4-scout-17b-instruct-v1:0} & \multirow{2}{*}{Open source} & 109 billion & \multirow{2}{*}{Meta} & \multirow{2}{*}{AWS Bedrock} \\
 &  & (17 billion active) &  &  \\
\multirow{2}{*}{meta.llama4-maverick-17b-instruct-v1:0} & \multirow{2}{*}{Open source} & 400 billion & \multirow{2}{*}{Meta} & \multirow{2}{*}{AWS Bedrock} \\
 &  & (17 billion active) &  &  \\
\botrule
\end{tabular*}

\footnotetext{Large Language Models Used with parameter counts (if known) and information about model version, creator and license}
\end{sidewaystable}
\begin{figure}[p!]
    \centering
    \includegraphics[width=\linewidth]{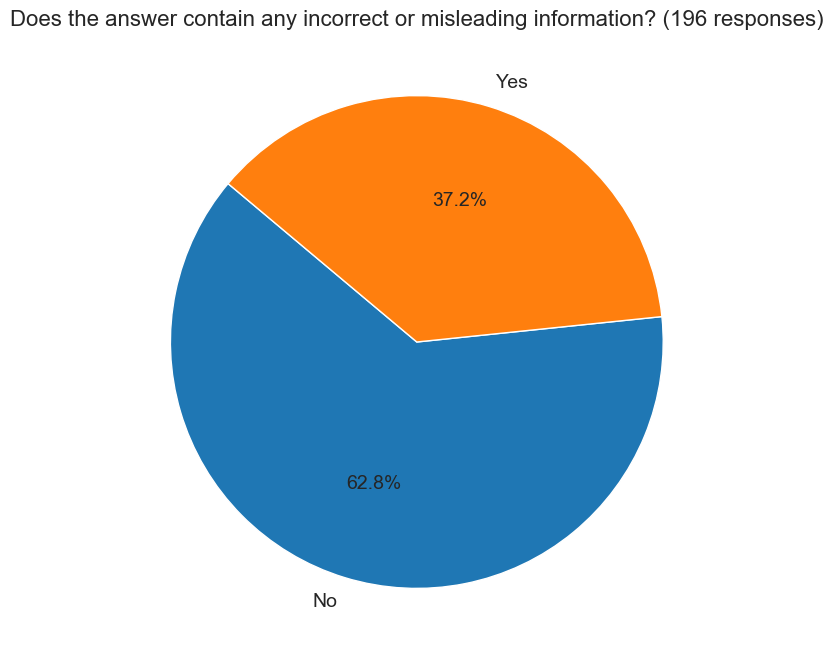}
    \caption{Response to the question ``Does the answer contain any incorrect or misleading information?"}
    \label{fig:pie_incorrect_misleading}
\end{figure}

\begin{figure}[p!]
    \centering
    \includegraphics[width=\linewidth]{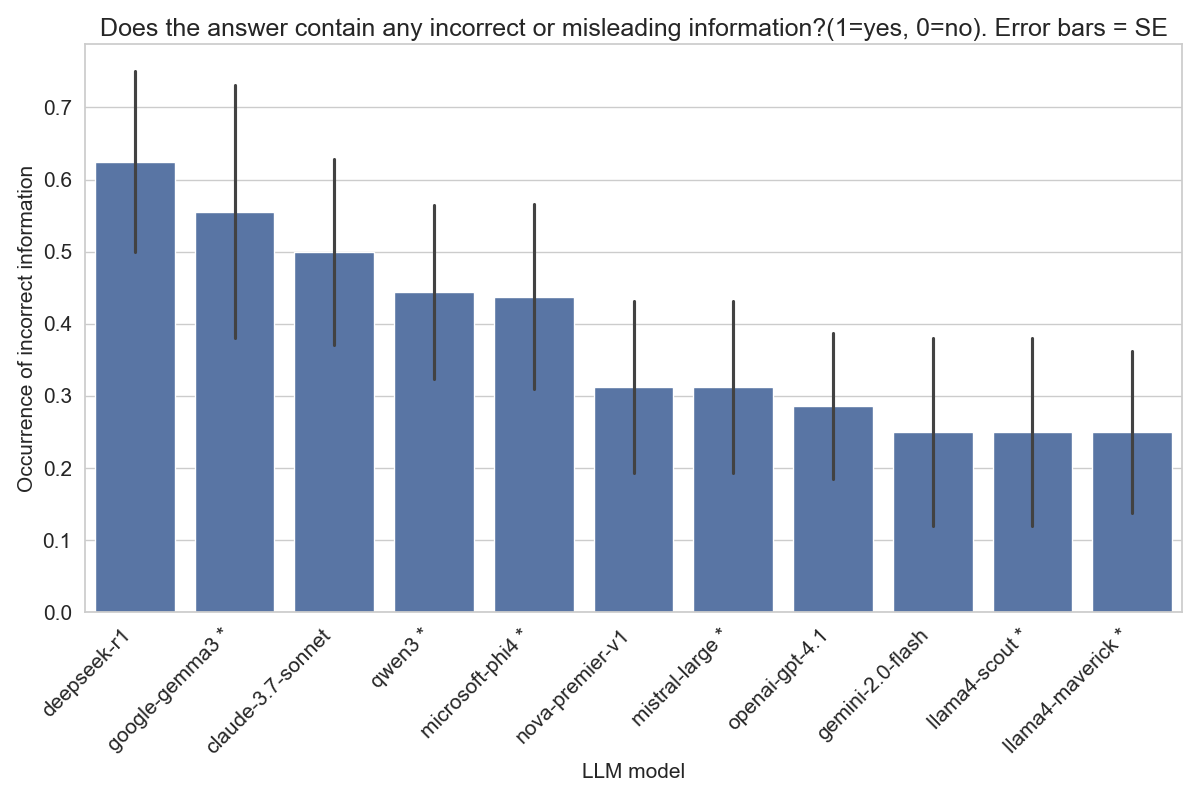}
    \caption{Response to the question ``Does the answer contain any incorrect or misleading information?" per LLM. Error bars indicate standard error. Asterisks indicate open-weight models.}
    \label{fig:bar_incorrect_per_llm}
\end{figure}
\begin{figure}[p!]
    \centering
    \includegraphics[width=\linewidth]{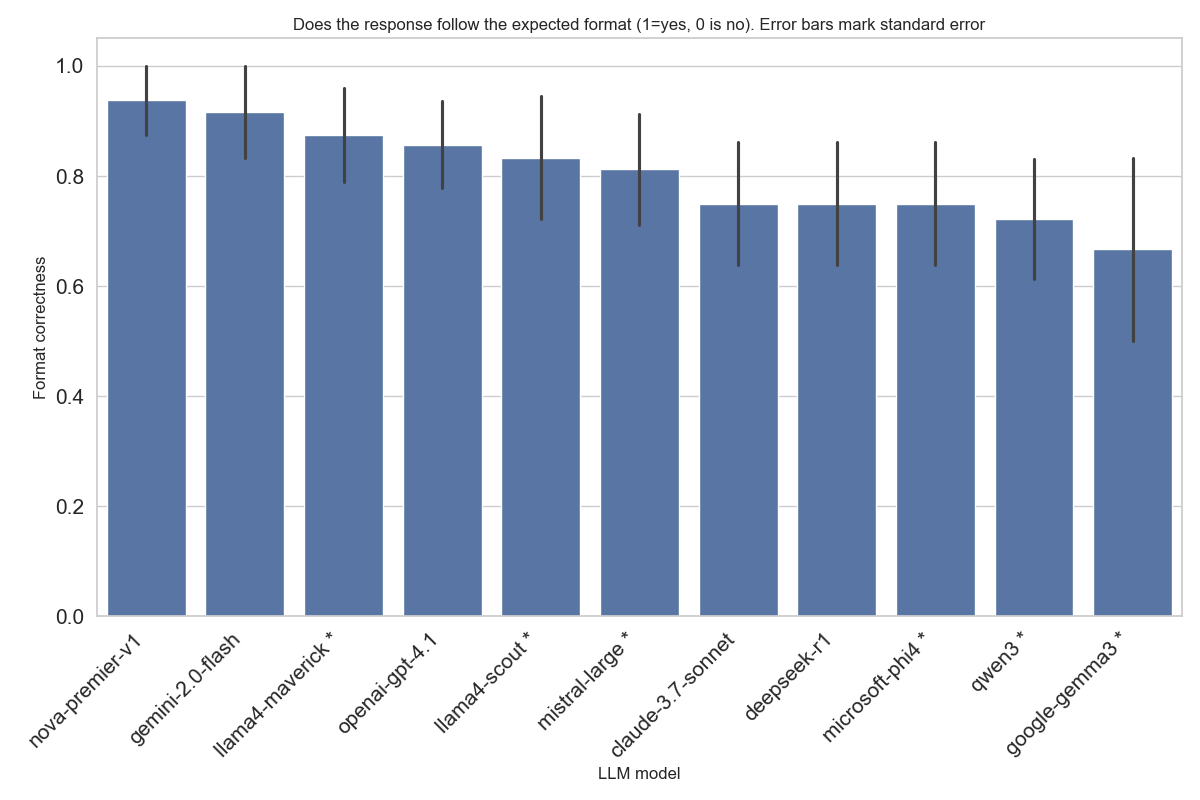}
    \caption{Bar chart showing expert rating of adherence to expected format across all models. Error bars indicate standard error. Asterisks indicate open-weight models.}
    \label{fig:format_all_models}
\end{figure}
\begin{figure}[p!]
    \centering
    \includegraphics[width=\linewidth]{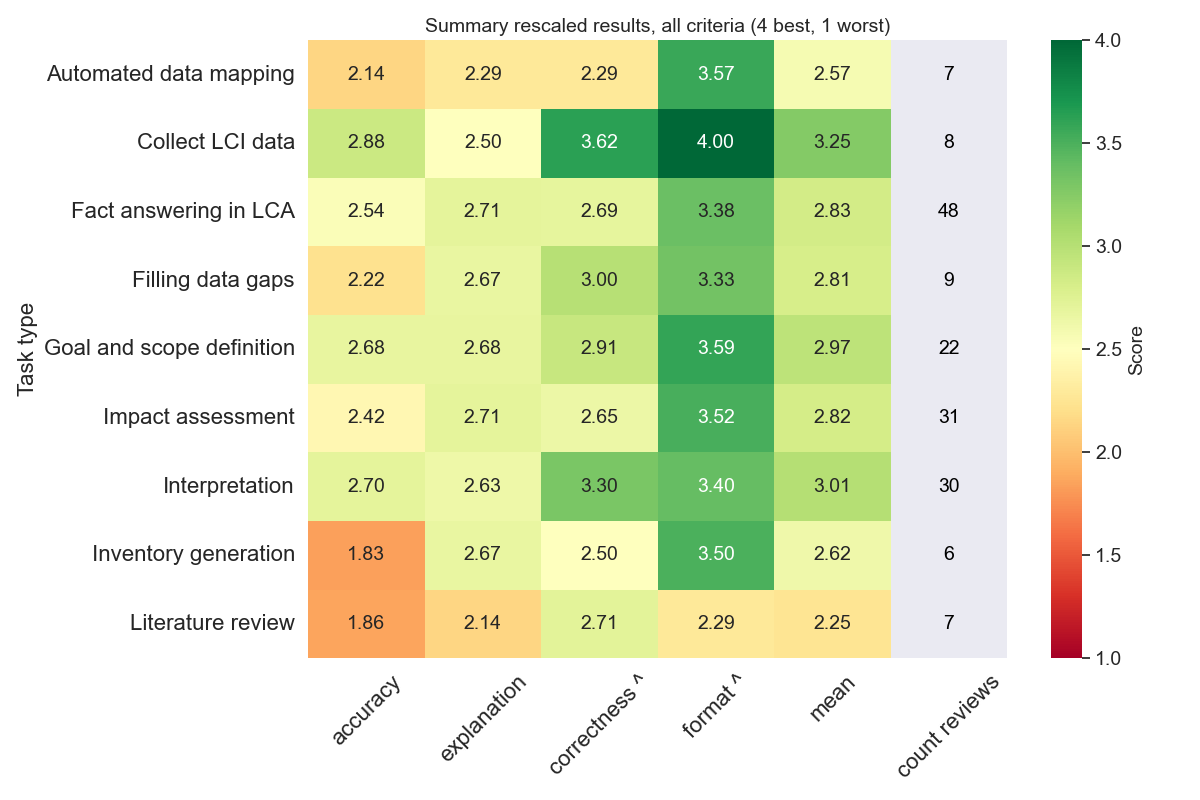}
    \caption{A heat map showing performance on criteria by task type, where 1 (red) is worst and 4 (green) is best. Scores for correctness and format adherence are scaled from the interval [0,1] to [1,4]. Correctness results are the inverted results from the binary question on inaccuracy (sec. \ref{correctness_calc}). The right-most column shows the number of expert reviews included. }
    \label{fig:heatmap_tasktype}
\end{figure}




\end{appendices}

\clearpage 
\bibliography{sn-bibliography}

\end{document}